\documentclass[letterpaper, 10 pt, conference]{ieeeconf}  

\IEEEoverridecommandlockouts                              

\usepackage{graphics} 
\usepackage{epsfig} 
\usepackage{amsmath} 
\usepackage[english]{babel}
\usepackage[utf8]{inputenc}
\usepackage{algorithm}
\usepackage{siunitx}
\usepackage{color}
\usepackage{colortbl}
\usepackage{comment}
\usepackage{caption}
\usepackage{booktabs}       
\usepackage[noend]{algpseudocode}
\def\etal{\emph{et al. }}
\title{\LARGE \bf
A Surface Geometry Model for LiDAR Depth Completion 
}

\author{Yiming Zhao, Lin Bai, Ziming Zhang and Xinming Huang*
\thanks{The authors are with Department of Electrical and Computer Engineering, Worcester Polytechnic Institute,
        Massachusetts 01609, USA.
        {\tt\small Email: yzhao7@wpi.edu}}%
}

\begin{document}

\maketitle
\thispagestyle{empty}
\pagestyle{empty}

\begin{abstract}

LiDAR depth completion is a task that predicts depth values for every pixel on the corresponding camera frame although only sparse LiDAR points are available. Most of the existing state-of-the-art solutions are based on deep neural networks, which need a large amount of data and heavy computations for training the models. In this letter, a novel non-learning depth completion method is proposed by exploiting the local surface geometry that is enhanced by an outlier removal algorithm. The proposed surface geometry model is inspired by the observation that most pixels with unknown depth have a nearby LiDAR point. Therefore, it is assumed those pixels share the same surface with the nearest LiDAR point, and their respective depth can be estimated as the nearest LiDAR depth value plus a residual error. The residual error is calculated by using a derived equation with several physical parameters as input, including the known camera intrinsic parameters, estimated normal vector, and offset distance on the image plane. The proposed method is further enhanced by an outlier removal algorithm that is designed to remove incorrectly mapped LiDAR points from occluded regions. On KITTI dataset, the proposed solution achieves the best error performance among all existing non-learning methods and is comparable to the best self-supervised learning method and some supervised learning methods. Moreover, since removing outlier points coming from occluded regions is a commonly existing problem in LiDAR-Camera systems, the proposed outlier removal algorithm is a general preprocessing step that is applicable to many robotic systems with both camera and LiDAR sensors. \footnote{https://github.com/placeforyiming/RAL\_Non\mbox{-}Learning\_DepthCompletion}

\end{abstract}

\section{INTRODUCTION}

LiDAR is playing an important role for modern mobile robots. Many LiDAR related perception tasks, such as LiDAR 3D detection \cite{wang2019pseudo} or point cloud semantic segmentation \cite{milioto2019rangenet++} attract many recent contributions. LiDAR-based depth completion is one of those tasks which aims to provide the dense depth map for the camera from the sparse depth generated by mapping LiDAR points on the image. Before the booming of LiDAR technology, depth completion is usually referred as a challenge to complete missing depth from depth sensors like Microsoft Kinect \cite{zhang2018deep}. The first LiDAR-based depth completion paper \cite{uhrig2017sparsity} proposed the unique sparsity challenge as LiDAR depth image only contains around 5\% values. The benchmark on KITTI LiDAR depth completion task has attracted many successive research works \cite{Geiger2012CVPR}.

\begin{figure}
\includegraphics[width=0.95\linewidth, height=60mm]{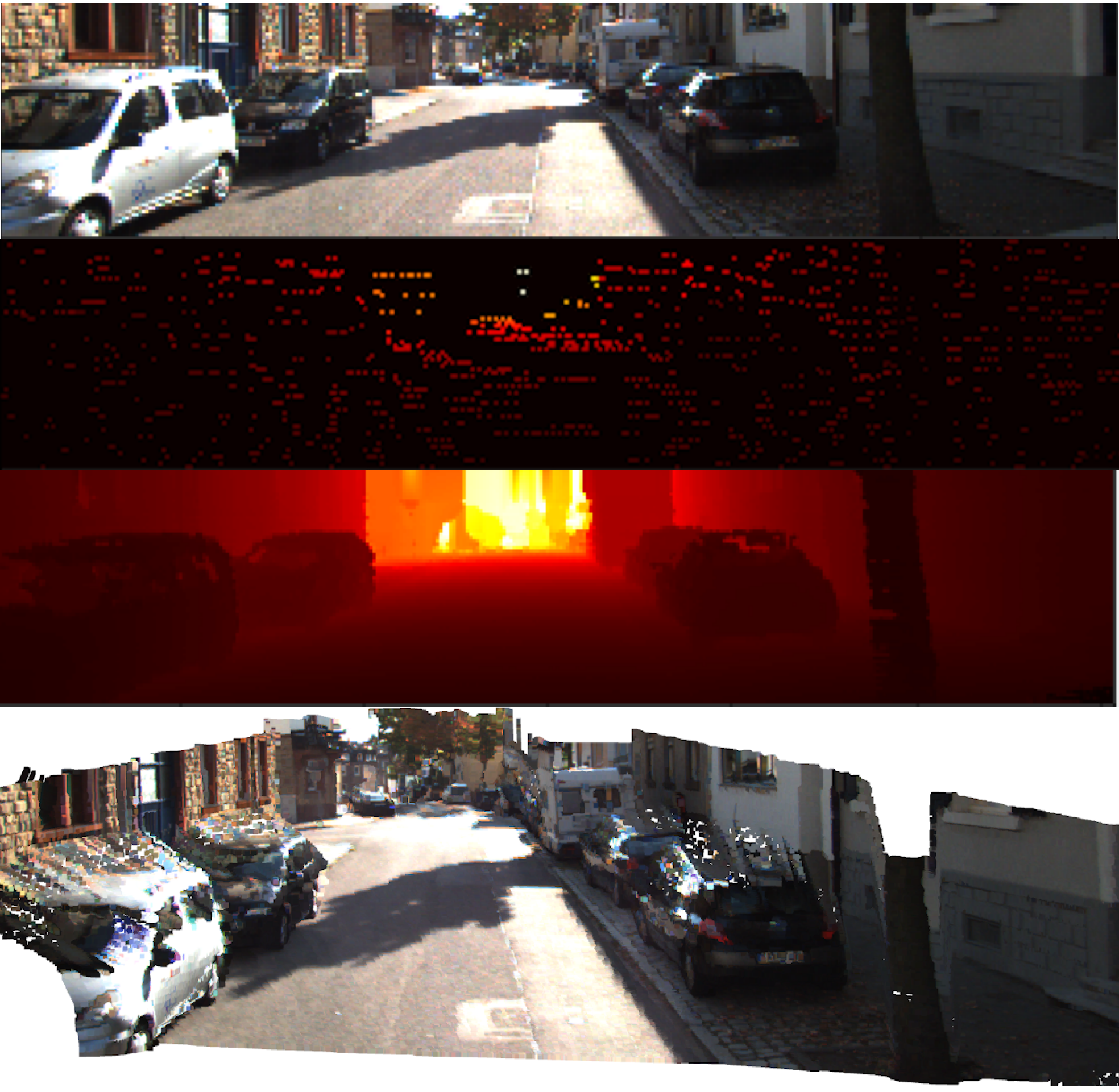}
    \caption{Illustration of the depth completion task. From top to bottom: RGB image, sparse LiDAR input, dense completed depth map, and colored 3D points in real world. This sample comes from the validation dataset in KITTI. }
    \label{fig:fisrt}
    \vspace{-7mm}
\end{figure}

A straightforward solution is using the neural network as a regression function trained with labels. Most of those papers are borrowing ideas from recently developed deep learning technologies, such as different post-process modules \cite{cheng2020cspn++}, or various network structures like encoder-decoder or hourglass \cite{ma2019self,li2020multi}. In contrast, this paper proposes a model-based explainable non-learning method to complete the depth map. In Fig. \ref{fig:fisrt}, we show one sample from our non-learning method. The generated depth map recovers the surroundings by embedding all the RGB values in the 3D world.


\textbf{Motivation} Many neural network methods have been proposed recently for the LiDAR depth completion task. Supervised learning requires labeled data, while self-supervised learning \cite{ma2019self} is able to train the network without labels. The label-free merit allows the method to easily adapt new environments by avoiding offline labeling work. However, a recent paper reports a simple non-learning method with classical image processing operators that achieves comparable performance with deep learning methods \cite{ku2018defense}. Since the non-learning solution shares the same label-free merit, we are intrigued to think what is the hidden mechanism behind this simple method with image processing operations like dilation.  

\textbf{Outlier Removal} Since LiDAR and camera are two sensors observing the world from different views, there will be some small occluded regions that are only observed by the LiDAR sensor but not observed by the camera. Mapping those points on image plane will create many incorrect depth values. To further evaluate this kind of error, we directly calculate the error between the raw sparse LiDAR and the ground truth on the KITTI validation set. The RMSE(root mean square error) is 1595mm that is even larger than the completed depth map generated by many models. This explains why dilation works as it replaces some of those erroneous values with the smallest value in the local patch. However, the dilation will also change values of many LiDAR points that are not occluded. To solve this challenge, we design a general outlier removal algorithm as a preprocessing step for the non-learning depth completion method.

\textbf{Local Surface Geometry} Besides the outlier removal, we also figure out why filling in the empty pixel with the smallest value in a local patch will not generate large errors. We identify the relationship between the empty pixel with its surrounding points is decided by the local surface geometry. Thus, we make an assumption that most empty pixels are likely sharing the same surface with their nearest point. Then, the empty depth value is estimated as an initial guess from the nearest value plus a residual determined by the direction of the local surface and the relative distance. In fact, our model is able to produce solid performance on KITTI, which demonstrates the effectiveness of the sharing local surface assumption. Note, the sharing surface assumption implies the point cloud should contain enough geometry information of objects, such as the commonly used 64 line LiDAR. We further discuss the model performance with various input sparsity in Section V.

\textbf{Contribution} Due to the space limitation, we prepare more supportive materials about the sharing surface assumption and outlier removal algorithm in the supplementary file. We summarize two major contributions of this paper as following:

\begin{itemize}
    \item \emph{We propose a general outlier removal algorithm to remove incorrect depth values projected on the image by points from occluded regions.} Considering the displacement of LiDAR and camera is usually small, the parallax effect created by the displacement is also smaller than most objects' size. This finding builds the intuition of our algorithm that those points with parallax effect should be identified as outliers. We further extend this intuition to three observations and successfully design an effective algorithm. The proposed outlier removal algorithm is parameter-free that only needs the image and LiDAR resolutions, i.e, 64 line or 32 line for LiDAR, $256 \times 256$ or $512 \times 512$ for image. Therefore, the algorithm is able to be directly applied in many LiDAR-camera systems as a preprocessing step.

    \item \emph{We propose a model-based explainable non-learning depth completion method and achieve good performance with several practical merits.} On KITTI, we achieve the best performance among all non-learning methods, and also achieve similar or better performance compared with all recent developed self-supervised learning methods. As an explainable model, it also clearly indicates the working condition that the point cloud should contain enough geometry information. Our model also shares all the merits of non-learning methods, such as label-free, robust to environment changing, and computational friendly.
    
\end{itemize}

\section{RELATED WORK}
\textbf{Classical methods for depth completion} Before the wide adoption of outdoor LiDAR for autonomous driving, depth completion was researched as a task to fill up the depth map generated from the RGB-D camera. Some classical methods were proposed to solve the image inpainting or depth quality improvement problem for RGB-D depth map \cite{lu2014depth,shen2013layer,park2014high,yu2013shading,silberman2012indoor}. The ratio of missing data from RGB-D depth map is roughly around 10\% to 30\%, which is different with sparsity from LiDAR.
Besides completing the generated depth map from the RGB-D camera, some researchers assumed only a few points have been matched by the stereo camera and complete this sparse input to dense by using optimization methods like compress sensing \cite{hawe2011dense,liu2015depth,ma2016sparse}. The natural difference between LiDAR sensors with those RGB-D sensors urges researchers to design new methods for outdoor LiDAR depth completion. Recently, Ku \etal \cite{ku2018defense} proposed a fast method for LiDAR-based depth completion with simple image processing operators, including dilation, Gaussian blur, etc. This simple method even gets the comparable result with some learning-based methods. We choose this method as one of our baseline comparisons.

\textbf{Supervised learning methods for depth completion} 
The recent success of deep neural networks on several computer vision tasks attracts researchers to solve the depth completion task by using networks. The supervised way treats the network as a regression model to directly output depth value. To handle sparse depth, various innovations are proposed in different aspects, like special designed layers \cite{ DBLP:conf/bmvc/EldesokeyFK18,8765412, eldesokey2020uncertainty},  encoder-decoder structure \cite{ma2019self}, fuse with semantic segmentation\cite{jaritz2018sparse,van2019sparse}, etc. Besides completion with LiDAR only, integrating image as a branch shows a relatively large improvement. Some works modeled RGB branch inspired by the use of surface normal \cite{qi2018geonet,yang2018unsupervised,qiu2019deeplidar}. Since LiDAR can capture the geometry information, the fusion of geometry information with image texture information was also investigated in some works \cite{xu2019depth,chen2019learning}. There are other ideas that also have been tried to solve this task.  Imran \etal \cite{imran2019depth} proposed a new depth representation named Depth Coefficients. Chen \etal \cite{cheng2020cspn++} designed a post-process module to further improve the model performance.

\begin{figure*}[t]
\includegraphics[width=\linewidth, height=60mm]{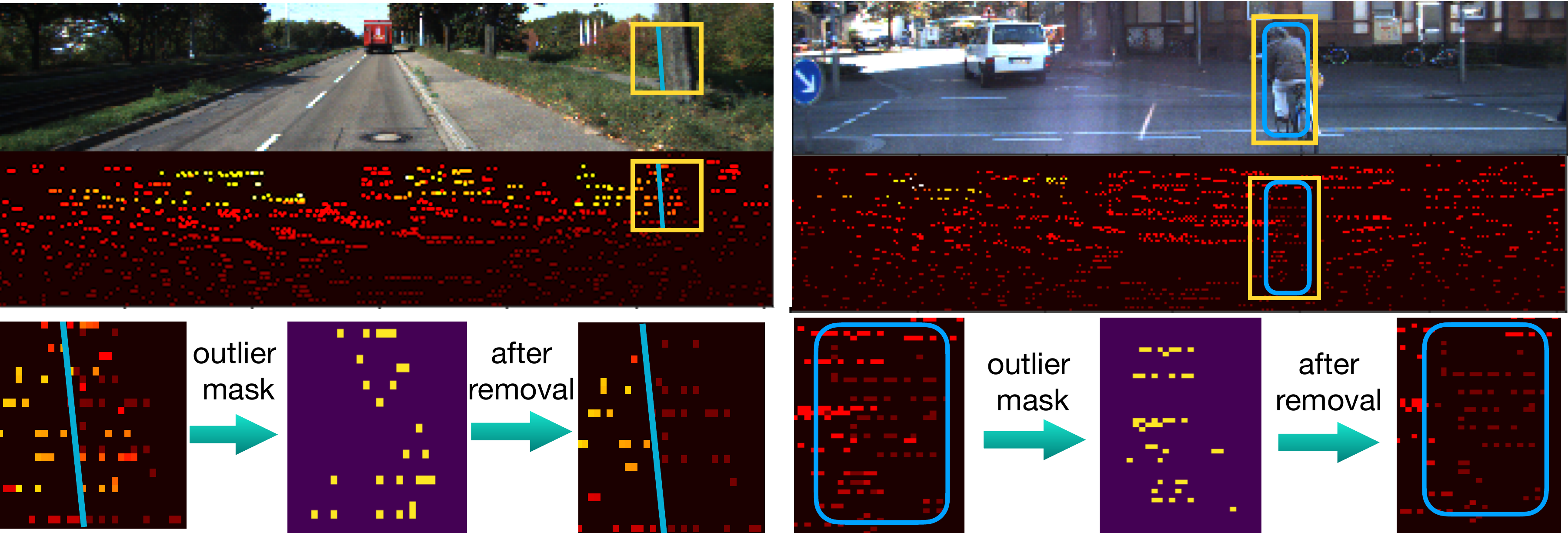}
\caption{Illustration of outliers and results of removal method. The darker color in LiDAR image means the depth is closer to the sensor, and black means no value. The displacement of LiDAR and camera will map some occluded points from the background on the foreground object. On the left sample, the blue line separates the left background and right foreground tree. On the right sample, the blue circle shows the foreground cyclist. Our method removes those incorrect points and keeps a clean edge for occluded regions.  
}
\vspace{-3mm}
\label{fig:h_1}
\end{figure*}

\textbf{Self-supervised learning methods for depth completion} 
Self-supervised learning can train the network without labels. Self-supervised depth prediction relies on the image warping and photometric loss to penalize the error \cite{zhou2017unsupervised,godard2017unsupervised}. Ma \etal \cite{ma2019self} extended the self-supervised depth prediction framework to depth completion by feeding the sparse points into the network and treating them as the ground truth for corresponding pixels. Moreover, Yang \etal \cite{yang2019dense} integrated their Conditional Prior Network to the self-supervised depth completion pipeline and got a better performance. Wong \cite{wong2020unsupervised} further considered pose consistency and geometric compatibility to improve performance. Yao \cite{yao2020discontinuous} proposed a binary anisotropic diffusion tensor to eliminate smoothness constraint at intended positions and directions. Those self-supervised methods do not need labels, which allows them to update weights online. This convenience is useful for many advanced situations like federated learning, domain adaptive learning, and user privacy protection.

\textbf{In summary} Self-supervised methods have label-free merit. On the contrary, supervised methods need labels but usually have better performance. Moreover, from the view of input modality, sensor fusion methods with both LiDAR and RGB have better performance, but LiDAR only methods are not sensitive to the change of lighting conditions, thus is stable in extreme cases like dark night. Our method is label-free and only needs LiDAR as input. We compare the performance with all label-free methods on the popular KITTI benchmark as well as show additional benefits on a night-time sample from the Waymo dataset qualitatively.   

\section{Methods}

\subsection{Outlier Removal}
Though camera and LiDAR are usually mounted close to each other, there is still a small displacement between them. Due to the slightly different views, the multi-planar parallax will create some occluded regions which are observed by LiDAR but should not be observed by camera. Mapping those points on image plane leads to a large error as those points are coming from far away background. To address this challenge, we propose a general outlier removal algorithm that can work on arbitrary image and LiDAR resolutions. In Fig. \ref{fig:h_1}, we show two examples that many points from the background are mapped onto foreground tree or cyclist. Our proposed solution will identify and remove those points and keep a clean boundary for objects.    

Our removal algorithm is based on three observations: 
\begin{itemize}
    \item \emph{Most outlier points will change the relative position with some nearby correct points as they are occluded by foreground objects.} Although there are some exceptions, for instance, the outlier is just mapped on the edge of the foreground, or the size of the foreground object is small. We argue most outliers will have this property. 
    \item \emph{Outlier points have relatively larger depth values compared with nearby correct depth values since they are coming from further background.} 
    \item \emph{Outlier points will create denser regions on image since they are mixed with points belong to the foreground.} 
    
\end{itemize}

\begin{figure}[t]
\includegraphics[width=0.9\linewidth, height=40mm]{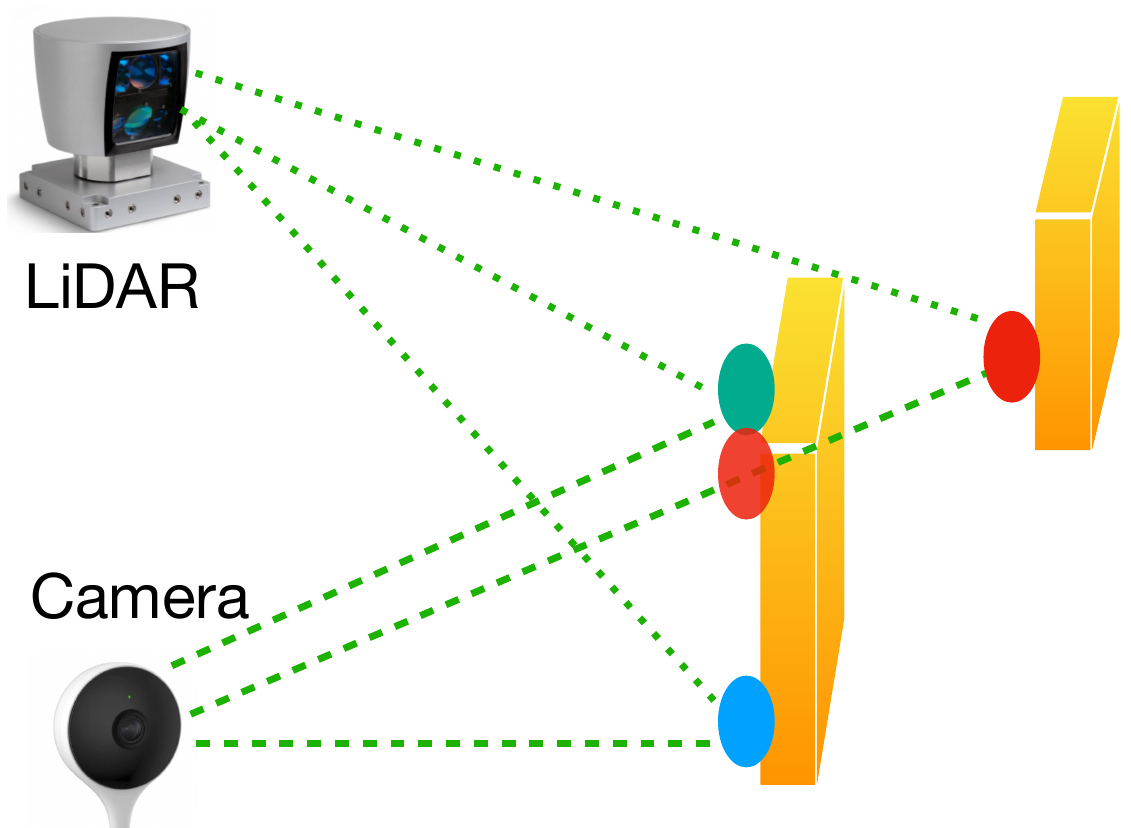}
    \caption{Illustration of our three observations. The red point is the outlier that should be removed. It is on the left of the green point in the LiDAR coordinate but on the right of the green point in the camera coordinate. The red point has a larger depth value than green point and also creates a relatively denser region with green point on image plane. }
    \label{fig:h_2}
    \vspace{-5mm}
\end{figure}

We further illustrate those three observations in Fig. \ref{fig:h_2}. They build the foundation of our outlier removal algorithm.  

Here, we assume that we have an image with the resolution $W\times H$, and a LiDAR with $L$ lines. There are $N$ points in total which are mapped on the image plane. Average distances between two points in two directions on image plane can be roughly estimated as $\frac{WL}{N}$ and $\frac{H}{L}$. Therefore, for a specific point $i$ on the image plane $(\mu_{i}, \nu_{i})$, we define a set of nearby points in a local region as:
$$S(i)=\{n| \forall n\in (1,..,N), |\mu_{j}-\mu_{i}|<\frac{WL}{N} \mbox{ and } |\nu_{j}-\nu_{i}|<\frac{H}{L}\}$$
In the LiDAR coordinate, we choose the spherical representation for the point $i$ as $(r_{i}, \theta_{i}, \phi_{i})$. The depth value on the image plane is defined as $z_{i}$. 

In our outlier removal algorithm, the $condition\_1$ and $condition\_2$ are indicators to check if the point $i$ meet our first and second observations, respectively. Our third observation suggests that $S(i)$ should contain some points if point $i$ is an outlier. The $\epsilon$ is a small value, and we choose $\epsilon= 1.0m $ in this paper.
\begin{algorithm}
\caption{Outlier Removal}\label{alg:euclid}
\begin{algorithmic}
\For{ i in (1,2,...,N)}
\For { j in S(i)}
    \State $condition\_1\gets False$, $condition\_2\gets False$
\If{$(\mu_{i}-\mu_{j})*(\theta_{i}-\theta_{j})<0$ \textbf{or}\\
\hspace{1.2cm}  $(\nu_{i}-\nu_{j})*(\phi_{i}-\phi_{j})<0$}
    \State $condition\_1\gets True$
\EndIf

\If{$z_{i}>z_{j}+\epsilon$} 
    \State $condition\_2\gets True$
\EndIf

\If{$condition\_1==True$ \textbf{and}\\
\hspace{1.2cm} $condition\_2==True$} 
    \State remove i\\
    \hspace{1.55cm}\textbf{break}
\EndIf
\EndFor
\EndFor
\end{algorithmic}
\end{algorithm}

\subsection{Surface Geometry Model for Depth Completion}

\begin{figure}[ht]
\includegraphics[width=1.0\linewidth]{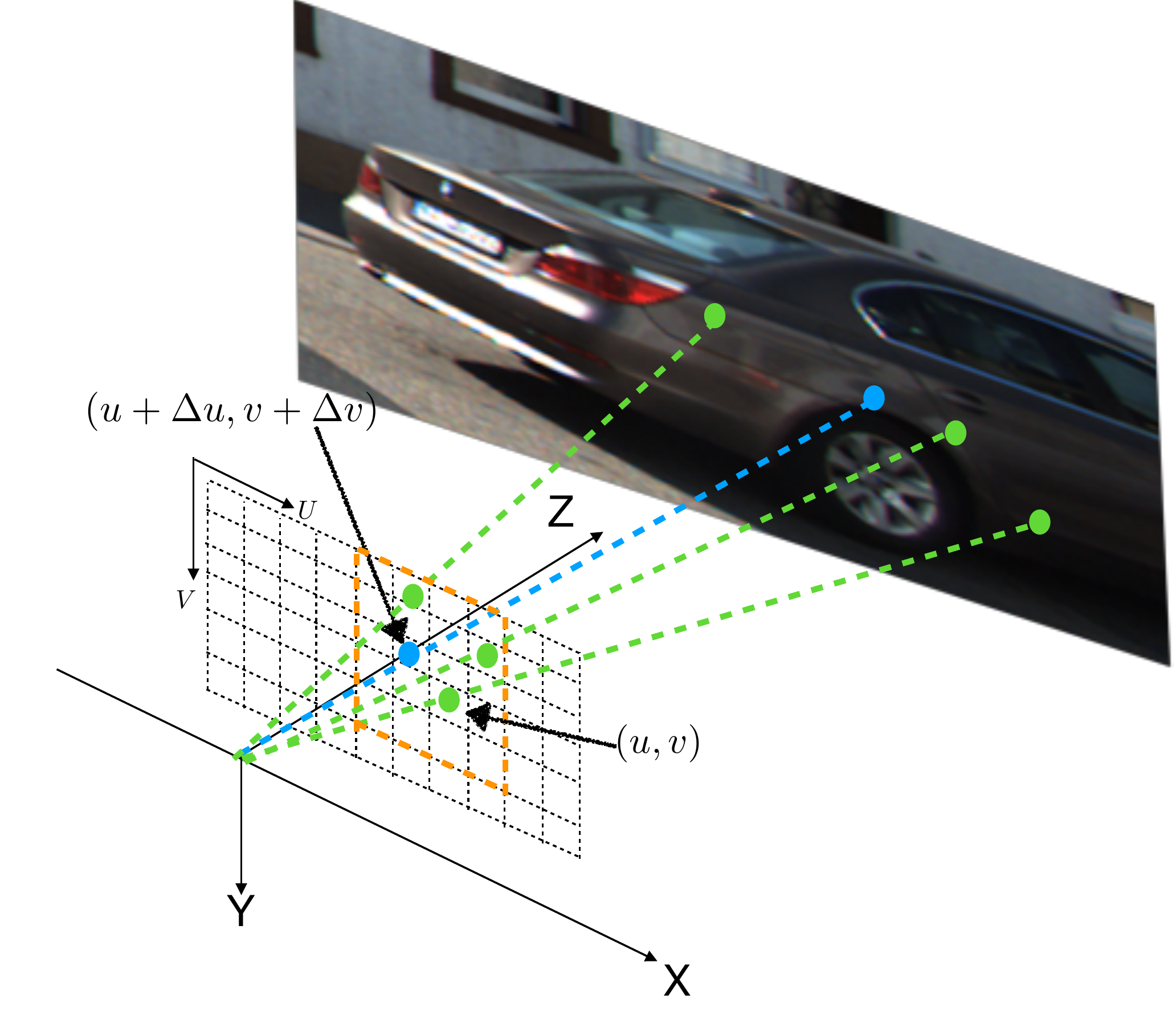}
    \caption{Illustration of our geometric model for depth completion. These empty pixels (green), which share the depth value from the same nearest signal (blue), are in a local patch and likely to belong to the same surface. }
    \label{fig:third}
    \vspace{-5mm}
\end{figure}

\textbf{Geometric Model}
Here we demonstrate a geometric model to complete the depth of empty pixel from its nearest pixel with value. We illustrate our intuition in Fig. \ref{fig:third}.  

As we can see, those empty green points share the same surface with the blue point in a small local patch. Let $[x,y,z]$ and $[x',y',z']$ be the 3D locations of one green point and the blue point. Further assuming that the intrinsic matrix of the camera is known and fixed as $\mathbf{\mathbf{K}}$, we can calculate the location of a 3D point $[x,y,z]$ in the camera coordinate as:
\begin{align}
    [x,y,z]^{T}=(z'+\Delta z)\mathbf{\mathbf{K}}^{-1}[u,v,1]^{T},
\end{align}

where $[u,v]$ and $[\Delta u, \Delta v]$ denote the 2D location and the offset between the two points in the depth image, respectively, $\Delta z$ denotes the difference between the estimated depth and the ground-truth, and $(\cdot)^{-1}, (\cdot)^T$ denote the matrix inverse and transpose operators, respectively. Similarly, we can compute $[x',y',z']$ as follows:
\begin{align}
    [x',y',z']^{T}=z'\mathbf{\mathbf{K}}^{-1}[u+\Delta u,v+\Delta v,1]^{T}.
\end{align}
Now considering the simplest pinhole model with focal length $[f_{u},f_{v}]$ and component point $[p_{u},p_{v}]$, we can get
\begin{align}
    & x=(z'+\Delta z)(u-p_{u})/f_{u}, & y=(z'+\Delta z)(v-p_{v})/f_{v}, \nonumber \\
    & x'=z'(u+\Delta u -p_{u})/f_{u}, & y'=z'(v+\Delta v-p_{v})/f_{v}. \nonumber
\end{align}
Since we know that $[u,v,z]$ and $[u+\Delta u, v+\Delta v,z']$ come from the same surface, we have the equation:
\begin{align}\label{eqn:abc}
    \mathbf{n}\cdot [x-x',y-y',\Delta z]^T=\mathbf{0},
\end{align}
where $\mathbf{n}$ is the normal vector of the surface.

Then, the depth value $z$ for empty pixel can be written as:
\begin{align}\label{eqn:a_2}
z=z'+\Delta z,
\end{align}
where
\begin{align}\label{eqn:abcd}
\Delta z =\frac{(\frac{z'\Delta \mu}{f_{\mu}},\frac{z'\Delta \nu}{f_{\nu}},0)\cdot \mathbf{n}}{(\frac{\mu - p_{\mu}}{f_{\mu}},\frac{\nu - p_{\nu}}{f_{\nu}},1)\cdot \mathbf{n}}
\end{align}

Eq. \ref{eqn:abcd} indicates the residual is decided by the nearest value itself, the offsets in image coordinate, the local surface normal, and camera parameters.

\textbf{Surface Normal} A common way to calculate the surface normal is fitting the local plane from the nearest neighbors of the query point \cite{rusu20113d}. However, those K-Nearest Neighbor based methods are too slow to process 64 line LiDAR point cloud which contains almost 20k points only in the front view. Fortunately, the special property of LiDAR sensor stimulates some more practical solutions. Following Badino's method \cite{badino2011fast}, we map LiDAR points from Cartesian coordinate system $(x, y, z)$ to spherical coordinate system $(r, \theta, \phi)$. Then, the normal vector is the derivative of function $r(\theta,\phi)$ in Cartesian coordinate, denoted as: 
\begin{align}\label{eqn:a_1}
\textbf{n} = \bigtriangledown_{(\hat{x},\hat{y},\hat{z})} r(\theta,\phi),
\end{align}
where $(\hat{x},\hat{y},\hat{z})$ is the unit vector in Cartesian coordinate.
 Eq. \ref{eqn:a_1} has solution:
\[ \textbf{n}=\left( \begin{array}{ccc}
\hat{x} &\hat{y}&\hat{z}
\end{array} \right)
R_{\theta,\phi}\left( \begin{array}{c}
1 \\
\frac{1}{rcos\phi}\partial r / \partial \theta  \\
\frac{1}{r}  \partial r / \partial \phi
\end{array} \right),
\]
where 

\[ R_{\theta,\phi}=\left( \begin{array}{ccc}
cos\theta&-sin\theta&0\\
sin\theta&cos\theta&0\\
0&0&1
\end{array} \right)
\left( \begin{array}{ccc}
cos\phi&0&-sin\phi\\
0&1&0\\
sin\phi&0&cos\phi
\end{array} \right).
\]

\begin{figure*}[t]
\includegraphics[width=\linewidth, height=60mm]{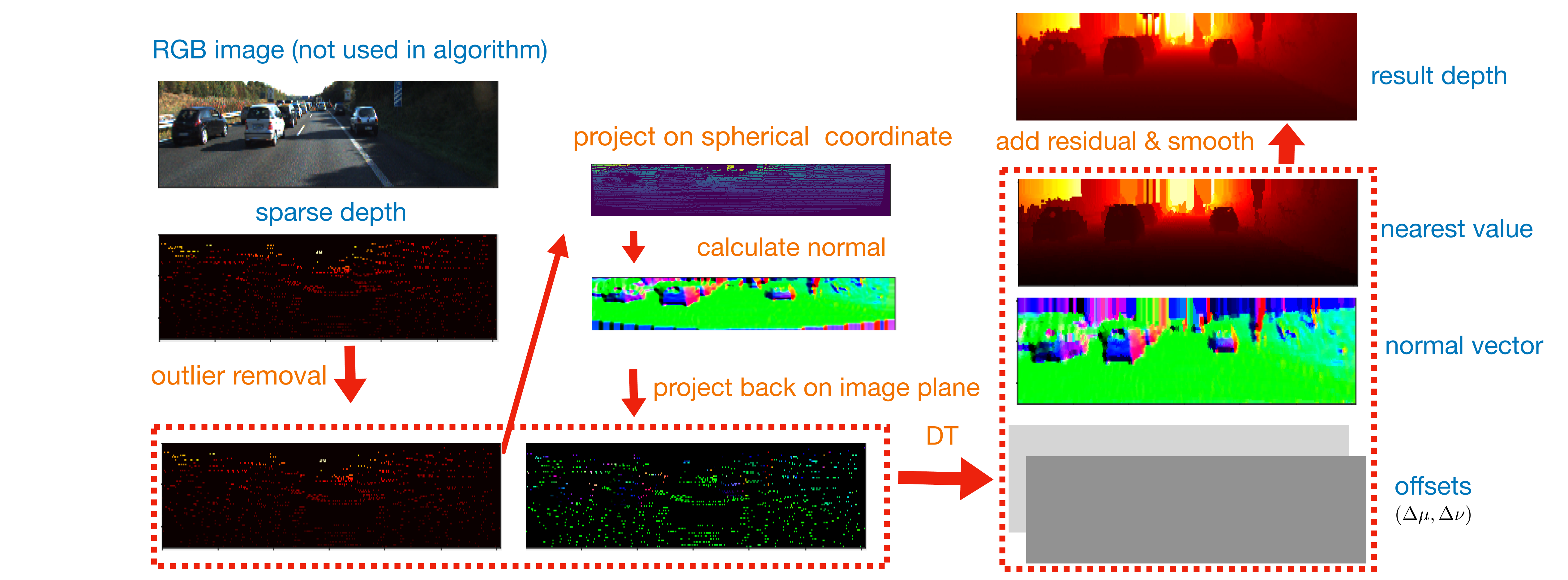}
\caption{Illustration of our non-learning depth completion pipeline based on local geometry model. Red arrows show the order of each step. The algorithm does not need RGB image, we put an RGB image here to show both the normal vector and final depth map are meaningful. We can see the visualized normal image is consistent with the RGB image, which also verifies the sharing surface assumption.
}
\vspace{-3mm}
\label{fig:hh}
\end{figure*} 

This method \cite{badino2011fast} provides a fast way to calculate normal just like edge detection. We add one extra step to fill in those empty pixels, which gives us smooth normal for all points. We show an example of the calculated normal vector in Fig. \ref{fig:hh}. The consistency between the normal image and the RGB supports the sharing surface assumption that most pixel shares the same surface with the nearest LiDAR point.

\textbf{Distance Transform \cite{Rosenfeld1966SequentialOI}} 
DT is an operator usually applied to binary images to generate distance maps (DM), which are grayscale images of the same size of the original images consisting of the distance as well as the offset to the closest available signal as each pixel value. For certain sampling functions, the complexity of DT is linear to the image size \cite{felzenszwalb2012distance}. This gives us a fast way to find $z'$ and $(\Delta \mu, \Delta \nu)$ needed for calculating $\Delta z$ in Eq. \ref{eqn:abcd}. It also distributes calculated normal vector to each empty pixel. 

\subsection{Pipeline Summary}

The whole pipeline of our non-learning method is summarized as four steps:

\textbf{Step 1:}
Project sparse points on spherical range image, calculate normal vector for each point, then project those points back on original camera image plane.

\textbf{Step 2:}
Run outlier removal algorithm to remove incorrect values on sparse depth image.

\textbf{Step 3:}
Run DT (distance transform) algorithm to prepare $z', \textbf{n}, (\Delta \mu,\Delta \nu)$ for each pixel on original image plane.

\textbf{Step 4:}
Apply Eq. \ref{eqn:a_2} and Eq. \ref{eqn:abcd} to calculate depth for each pixel. Then, smooth the result with Gaussian filter.

We visualize each step of our solution in Fig. \ref{fig:hh}.

\section{EXPERIMENTS}

\subsection{Dataset}

 We conduct our experiments on the KITTI dataset. The KITTI dataset is a benchmark and collection of sparse LiDAR scans and corresponding semi-dense depth maps which serve as ground-truth. The dataset consists of 85,898 training samples, 1,000 validation samples and 1,000 test samples. The ground-truth for test dataset is invisible, people need to submit the predicted test results onto the server to get the metric scores. Besides the numerical evaluation and

On KITTI, four evaluation metrics are used to evaluate model performance: 
\begin{itemize}
    \item \textit{RMSE}: Root mean squared error [mm]
     \item \textit{MAE}: Mean absolute error [mm]
      \item \textit{iRMSE}: Root mean squared error of the inverse depth [1/km]
       \item \textit{iMAE}: Mean absolute error of the inverse depth [1/km]
\end{itemize}

\subsection{Outlier Removal}

The error of sparse depth map has been reported by some works \cite{uhrig2017sparsity,qiu2019deeplidar}, here we evaluate the error quantitatively. We calculate the error of samples on the validation set for those pixels which have values in both sparse input depth and ground truth. The error is surprisingly large reported in the last row of Table \ref{tab:table1}. After removing those outliers, the error of sparse input has been significantly reduced.

\begin{table}[ht]
  \begin{center}
    \caption{The raw input error on KITTI's validation set.}
    \label{tab:table1}
        \setlength{\tabcolsep}{5pt} 
    \renewcommand{\arraystretch}{1.5} 

    \begin{tabular}{c|c|c|c|c|c}
      \hline
& {RMSE} & {MAE} & {iRMSE} & {iMAE} &{keep ratio} \\
       \hline
       original input& 1595.24 &202.16 & 5.00 & 0.75& 100.0\%\\ 
              \hline
       after removal& 662.87 &81.68 & 2.08 & 0.35& 94.0\%\\ 
\hline
    \end{tabular}
  \end{center}
  \vspace{-4mm}
\end{table}

Our outlier removal algorithm is also beneficial to other methods. For instance, self-supervised depth completion methods utilize raw sparse input as ground truth for corresponding pixels. Simply removing outliers will improve the performance, shown in Table \ref{tab:table2} by using SparseToDense \cite{ma2019self} as an example.

\begin{table}[ht]
  \begin{center}
    \caption{Outlier removal algorithm improves the performance of self-supervised solution on KITTI's validation set.}
    \label{tab:table2}
        \setlength{\tabcolsep}{5pt} 
    \renewcommand{\arraystretch}{1.5} 

    \begin{tabular}{c|c|c|c|c}
      \hline
& {RMSE} & {MAE} & {iRMSE} & {iMAE}  \\
       \hline
       SparseToDense \cite{ma2019self}& 1343.33 &358.57 & 4.27&1.63 \\ 
              \hline
       SparseToDense + removal& 1321.42 &331.19 & 4.07 & 1.60\\ 
       \hline
    \end{tabular}
  \end{center}
  \vspace{-4mm}
\end{table}

\subsection{The Performance of Surface Geometry Model}

\begin{table}[ht]
   \vspace{-1mm}
  \begin{center}
    \caption{Comparison on KITTI's test leaderboard. Red indicates the best, and blue indicates the runner-up. We use d to represent LiDAR. The bottom two methods are state-of-the-art performance of supervised models with single LiDAR as input and with RGB + LiDAR as input respectively. }
    \label{tab:table4}
    \setlength{\tabcolsep}{4pt} 
    \renewcommand{\arraystretch}{1.5} 
    \begin{tabular}{c|c|c|c|c}
      \hline
       Methods & Input & Type &{RMSE} &  {MAE}    \\
       \hline
         SparseToDense \cite{ma2019self}&RGB + d & self-supervised & 1299.85 & 	350.32  \\
              \hline
      DDP \cite{yang2019dense} &RGB + d&self-supervised   &  1263.19  & 343.46  \\
              \hline
    VOICE D\cite{wong2020unsupervised}  &RGB + d&self-supervised  &  \textcolor{red}{1169.97} & 299.41   \\
              \hline
            B-ADT\cite{yao2020discontinuous}  &RGB + d&self-supervised  &  1480.36 & \textcolor{blue}{298.72}  \\
              \hline
       SGDU \cite{schneider2016semantically} &RGB + d &non-learning & 2020.00 & 570.00 \\
                   \hline
       Bilateral \cite{barron2016fast} &RGB + d &non-learning & 1750.00 & 520.00\\

               \hline
     IP\_Basic \cite{ku2018defense} & single d &non-learning &   1288.46 	&302.60 \\
               \hline
      Ours   & single d &non-learning &   \textcolor{blue}{1239.84} &\textcolor{red}{298.30} \\ 	
               \hline
               \hline
                      SparseConv \cite{uhrig2017sparsity} &single d &supervised &1601.33 &481.27 \\
       \hline
              ADNN \cite{chodosh2018deep} &single d &supervised &1325.37 & 439.48  \\

       \hline
              NConv-CNN \cite{8765412} &single d &supervised &1268.22 &360.28  \\
       \hline
       
    \multicolumn{5}{c}{State-of-the-art on KITTI leaderboard}\\
                \hline
           IR\_L2 \cite{lu2020depth}   &single  d&supervised &  901.43 &	292.36 \\
               \hline
               \hline
          NLSPN \cite{park2020non}  &RGB + d&supervised&741.68 &199.59  	\\
               \hline

    \end{tabular}
  \end{center}
   \vspace{-4mm}
\end{table}

As a non-learning solution, our surface geometry model has the advantage of directly working in new environments without extra labeling work. This is also the major contribution of recent developed self-supervised depth completion methods \cite{yao2020discontinuous, wong2020unsupervised}. So we compare the performance of our method with all non-learning methods and self-supervised methods on KITTI's test leaderboard in Table \ref{tab:table4}. We achieve the best MAE and the second best RMSE among all non-learning and self-supervised learning methods. Our method even can outperform some early-stage supervised solutions. Recent developed supervised models are prone to use a large network with 10 million or even 100 million parameters \cite{lu2020depth,qiu2019deeplidar,park2020non}, thus have better performance. Besides, we also list two state-of-the-art supervised methods in Table \ref{tab:table4}. 

To complete the depth, two different input modality settings can be used. Sensor fusion models with RGB and LiDAR as input usually have better performance. However, it is still questionable if the method works under various lighting conditions. On the contrary, models with single LiDAR as input will not be affected by the change of lighting conditions. The method proposed in this paper has such advantage as it only relies on geometry information.

\subsection{Ablation Study}

\textbf{Effectiveness of Outlier Removal} To better evaluate our outlier removal algorithm, we show the model performance on KITTI validation set in Table \ref{tab:table5}. It is clear to see removing outliers significantly improve the performance of our model. From both Table \ref{tab:table1} and Table \ref{tab:table5}, our outlier removal method should be considered as a general preprocessing step for many camera-LiDAR sensor fusion tasks.

\begin{table}[ht]
   \vspace{-1mm}
  \begin{center}
    \caption{Verify the effectiveness of our outlier removal on KITTI's validation dataset.}
    \label{tab:table5}
    \setlength{\tabcolsep}{6pt} 
    \renewcommand{\arraystretch}{1.5} 
    \begin{tabular}{c|c|c|c|c}
      \hline
         &{RMSE} &  {MAE}  & {iRMSE} & {iMAE}  \\
               \hline
        Ours w/o Outlier Removal & 1629.38 & 400.09 & 6.31 & 1.77\\
              \hline
         Ours w Outlier Removal  & 1319.12 & 301.45 & 4.13 & 1.24\\
              \hline
    \end{tabular}
  \end{center}
   \vspace{-4mm}
\end{table}

\textbf{Effectiveness of Each Step} Table \ref{tab:table5} illustrates the effectiveness of our outlier removal module. To further clarify the benefit of each step, in Table \ref{tab:table6}, we show the change of RMSE(root mean square error), MAE(mean absolute error) and density after each step. We can see every step brings some improvements, and surface geometry connects all those steps together as an explainable non-learning model. The first two rows are the same as the Table \ref{tab:table1}, one may check that table for more details. 

\begin{table}[ht]
   \vspace{-1mm}
  \begin{center}
    \caption{Verify the effectiveness of each step in our method on KITTI's validation dataset. }
    \label{tab:table6}
    \setlength{\tabcolsep}{6pt} 
    \renewcommand{\arraystretch}{1.2} 
    \begin{tabular}{|c|c|c|c|}
      \hline
       Step & {RMSE}  & {MAE} & Density(\%)  \\
                       \hline
        original input & 1595.24 &202.16  & 4.37\\
                       \hline
        + outlier removal & 662.87 &81.68 & 4.11\\
                       \hline
        + distance transform & 1550.22 & 349.88 & 100.00\\
                       \hline
        + residual  &1526.63 & 318.12 & 100.00\\
                      \hline
        + smooth & 1319.12 & 301.45 & 100.00\\
               \hline
    \end{tabular}
  \end{center}
     \vspace{-4mm}
\end{table}

\section{Discussion} 

\subsection{Input Sparsity and Model Performance}

\begin{figure}[ht]
\includegraphics[width=1.0\linewidth, height=70mm]{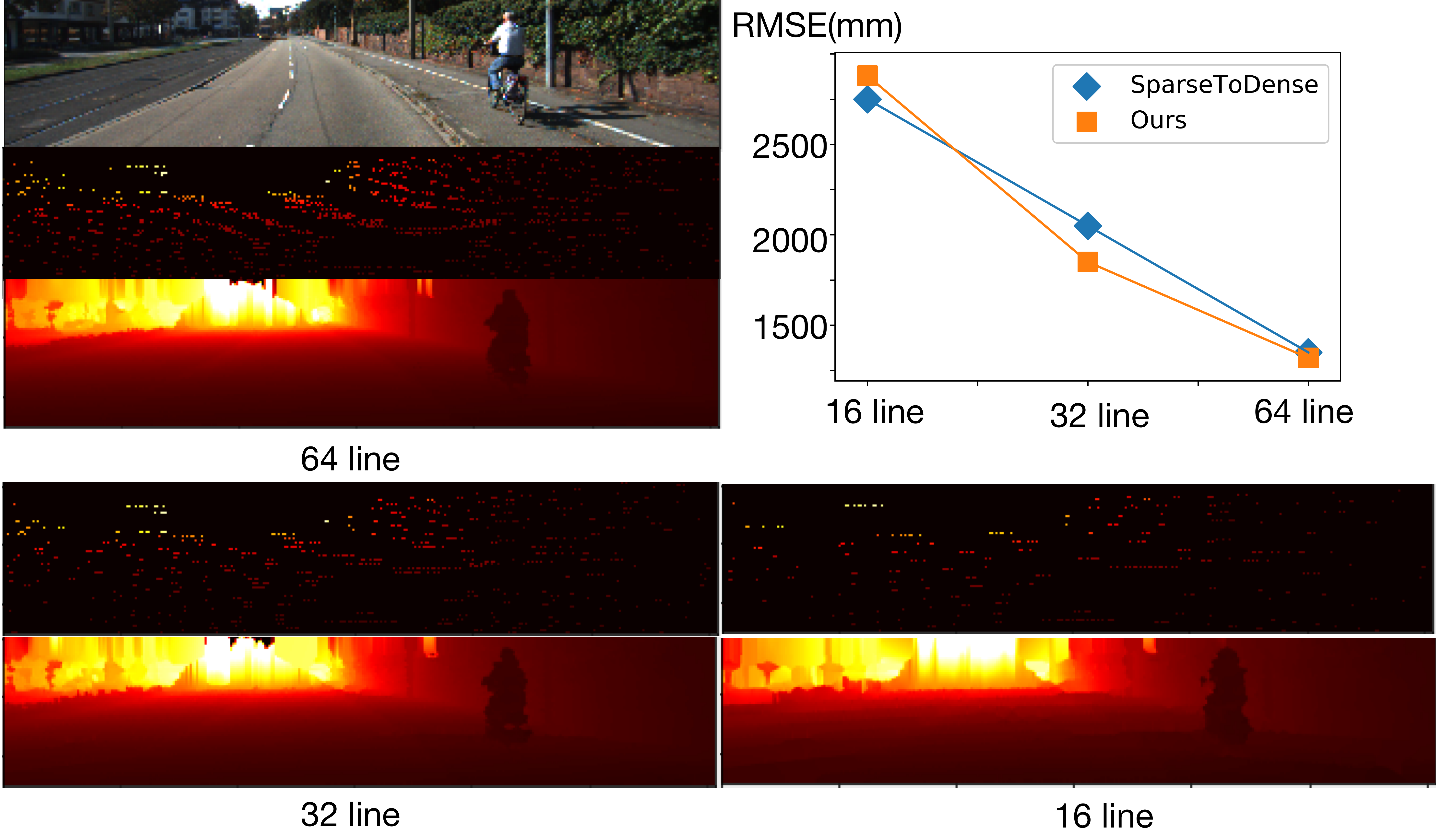}
    \caption{Compare model performance with self-supervised learning on three different density input: 64 line, simulated 32 line and simulated 16 line. }
    \label{fig:six}
    \vspace{-2mm}
\end{figure}

\begin{figure*}[t]
\includegraphics[width=0.99\linewidth, height=60mm]{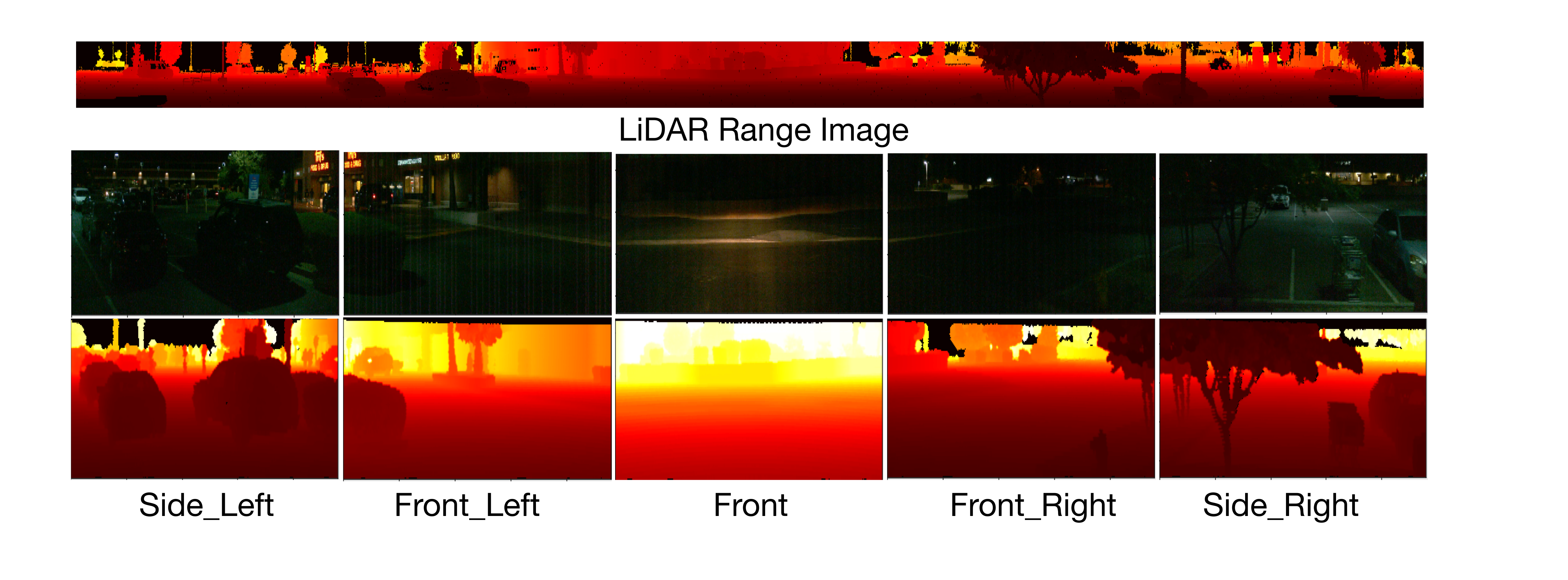}
\caption{Illustrate the ability of our method to provide depth maps for multi-cameras at night. This data sample is picked from the Waymo dataset.}
\vspace{-5mm}
\label{fig:waymo}
\end{figure*} 

The sharing surface assumption implies the point cloud should contain geometry information of objects, which is the case for commonly used 64 line LiDAR. It is worth to mention that recent papers even demonstrate a good result of panoptic segmentation on 64 line LiDAR only \cite{milioto2019rangenet++}. Here we discuss the performance of our method with sparser LiDAR density. By following the settings in SparseToDense \cite{ma2019self}, we group the 64 line LiDAR to downsample it as simulations of 32 line and 16 line LiDAR. We compare the performance of our model with the self-supervised results of SparseToDense \cite{ma2019self} in Fig. \ref{fig:six}. Our solution outperforms the baseline with 32 line and 64 line cases, but the self-supervised baseline starts to take the lead with 16 line case. This phenomenon verifies the condition of our algorithm that the point cloud can not be too sparse otherwise it may lose the geometry information. We also display the three different density LiDAR raw inputs and completed maps in Fig. \ref{fig:six}.

\subsection{Other Applications}

On KITTI's depth completion benchmark, we conduct the numerical evaluation and compare the performance with other depth completion methods. As a non-learning solution working on LiDAR only, our method has several benefits beyond numerical evaluation. First, our model is able to support driving under various lighting conditions, such as night-time driving. Second, owing to the label-free merit, our model is able to easily provide depth maps for cameras facing different directions. To further show those benefits, we give an example of our method on the Waymo dataset \cite{sun2020scalability}. As a recently published driving dataset, the Waymo dataset has more versatile driving scenarios. However, there is no depth completion benchmark on it which prohibits us to quantitatively evaluate our surface model. Instead, we use it to qualitatively demonstrate extra benefits of our solution in Fig. \ref{fig:waymo}. Based on the point cloud from the top LiDAR sensor, our method easily provides dense depth maps for five cameras mounted around the car at night. 



\section{CONCLUSIONS}
In this letter, we develop a novel non-learning depth completion method based on surface geometry. The method achieves the best MAE and the second-best RMSE compared with all label-free methods, including all non-learning and self-supervised learning methods on KITTI's leader board. This proposed algorithm is computationally efficient and can be used in any environment, therefore it is truly a practical solution to LiDAR depth completion task. Moreover, several individual modules, such as outlier removal or distributing normal vectors on the image plane, may also be beneficial to other robotic applications with LiDAR and camera sensors.





\section*{APPENDIX}

\subsection{Part 1. Statistical Information} 

\begin{figure}[ht]
\includegraphics[width=\linewidth, height=50mm]{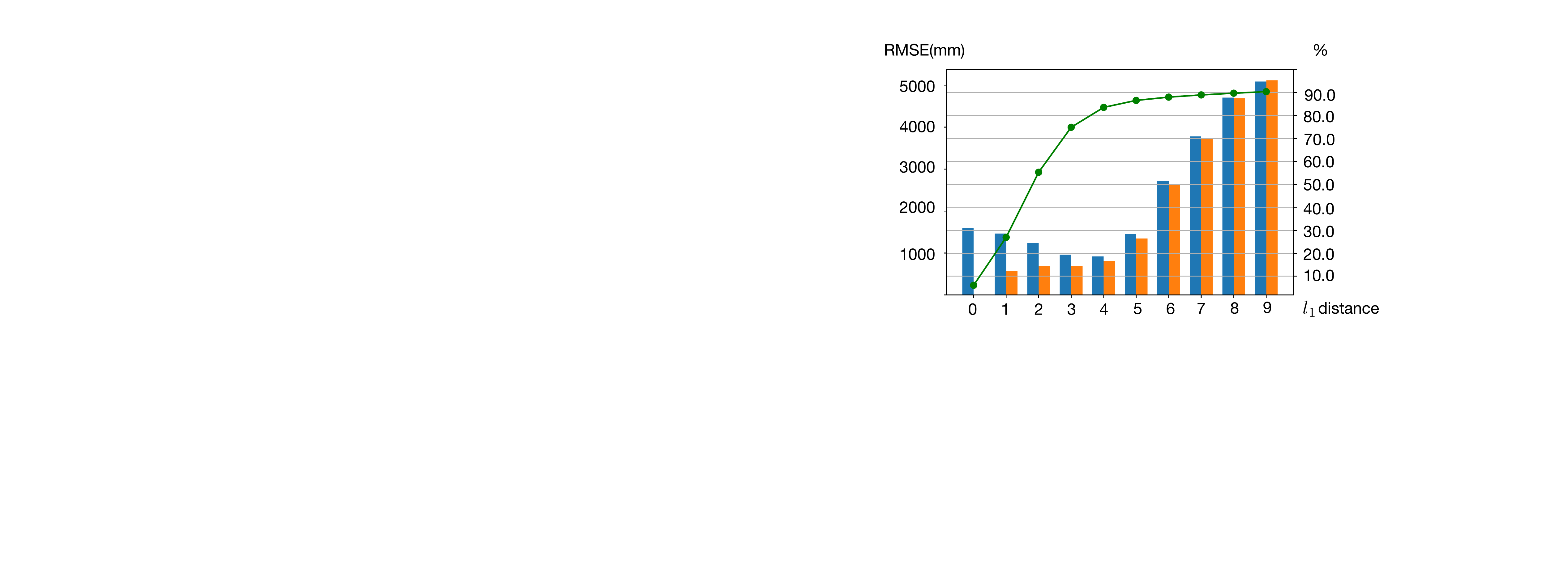}
\caption{ Statistical analysis on KITTI validation set. The \textbf{green} line is the percentage of the pixel that has a certain $l_1$ distance to the nearest point. The \textbf{blue} bar is the error between the nearest initial with ground truth. For the \textbf{orange} bar, we replace the input value with the ground truth, then calculate the nearest initial error again.}
\vspace{-3mm}
\label{fig:s_hh}
\end{figure}

Here we investigate the statistical information on KITTI validation set in Fig. \ref{fig:s_hh}. For each pixel on sparse input image plane, we calculate the $l_1$ distance to their nearest point, if the pixel itself has value, then the $l_1$ distance is 0. The y-axis is the percentage of the pixels that have the certain $l_1$ distance. We can see most pixels have a small distance to the nearest neighbor. Moreover, we explore the case that using the nearest value as the initial guess of empty pixels shown in blue bar. To further explore the effect of outliers, we replace the value of sparse input with the corresponding ground truth, then calculate the nearest initial again. We show this as the orange bar.
\begin{figure*}[ht]
\includegraphics[width=\linewidth, height=70mm]{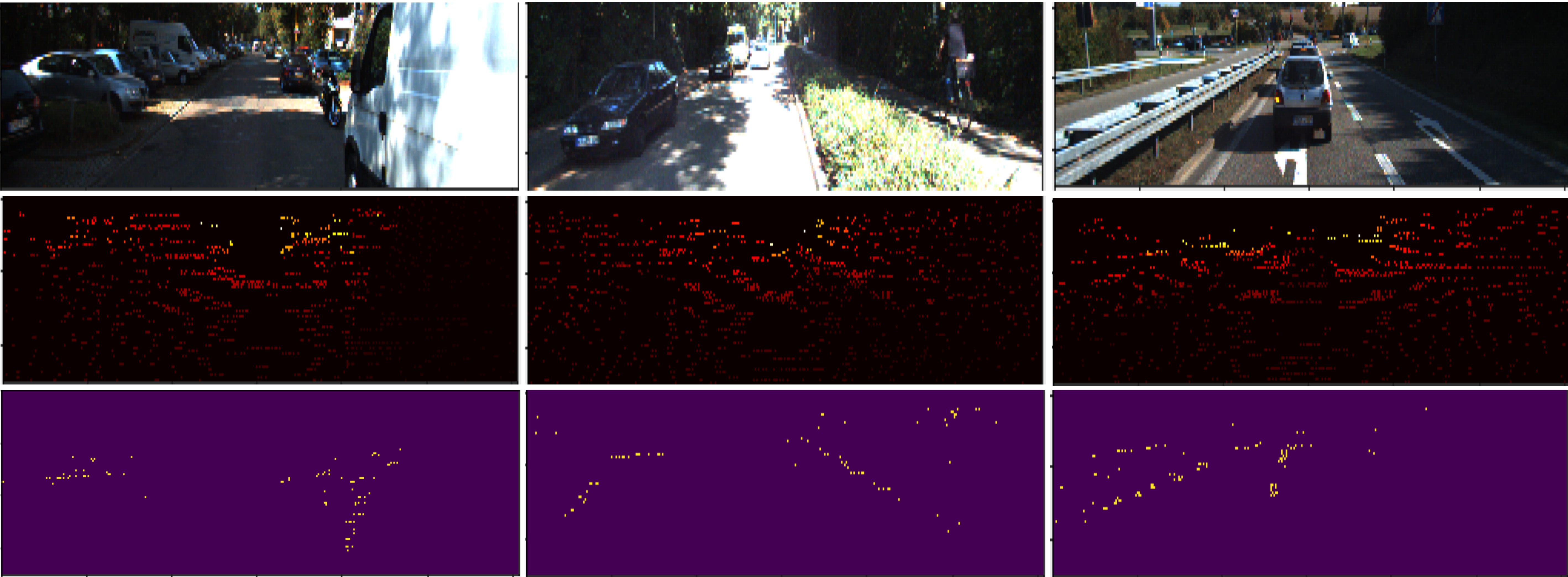}
\caption{ From top to bottom: RGB image, raw LiDAR sparse depth map, identified outlier mask.}
\vspace{-3mm}
\label{fig:s_hhhh}
\end{figure*} 
We can see several points from the statistical information:\\
a. Most empty pixels have a close nearest point. \\
b. Using the value of close nearest point as the initial guess has a small error.\\
c. Removing outliers will significantly reduce the nearest initial error. 

All those points support our sharing surface assumption, local surface geometry model and the importance of outlier removal algorithm.

\subsection{Part 2. Sharing Surface Assumption and Panoptic Segmentation}

\begin{figure}[ht]
\includegraphics[width=\linewidth, height=70mm]{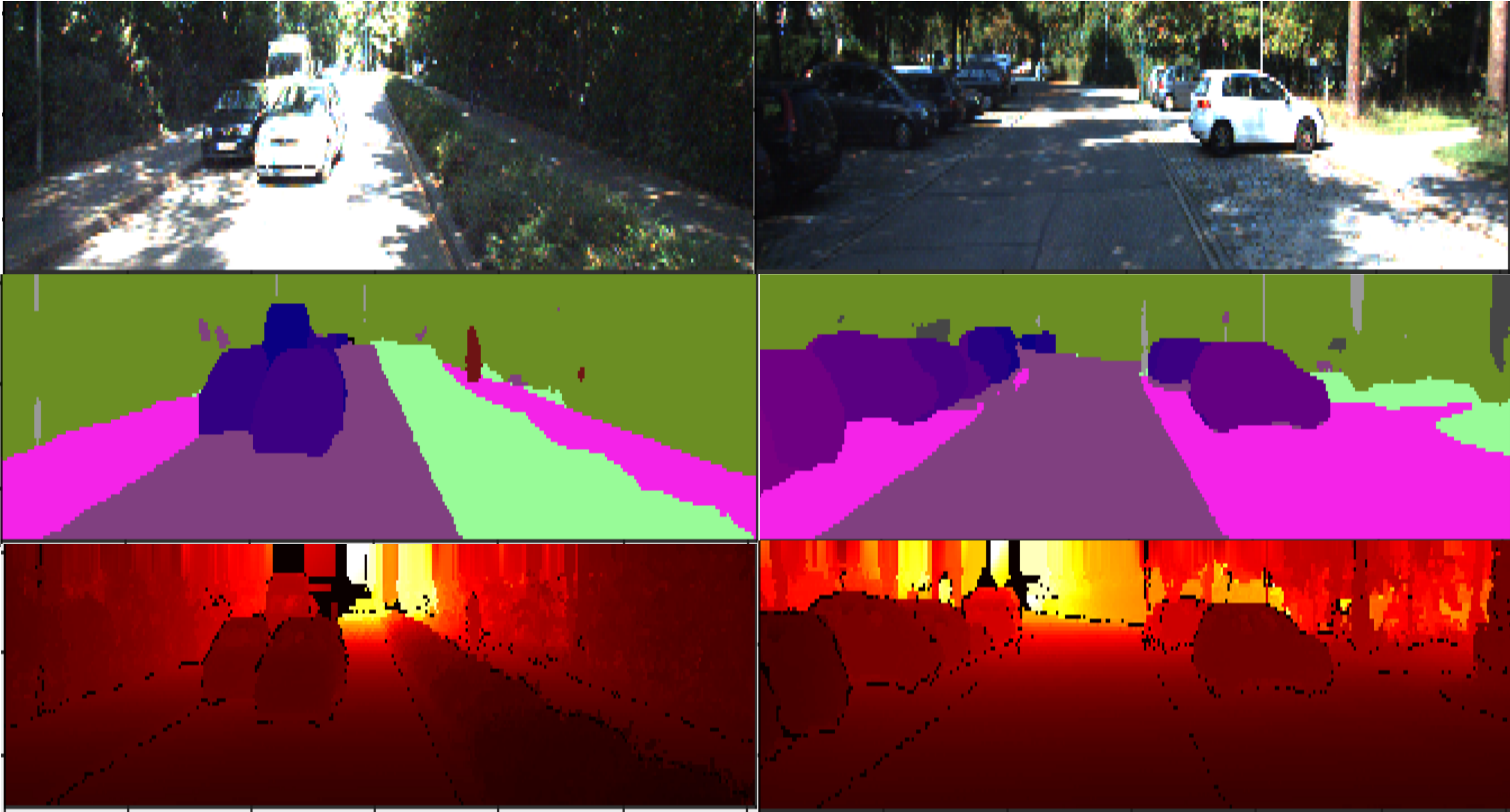}
\caption{ From top to bottom: RGB image, panoptic segmentation, completed depth map. Those black pixels on the depth map are coming from different objects determined by the panoptic mask.}
\vspace{-3mm}
\label{fig:s_hhh}
\end{figure}  

Due to the lack of normal ground truth, it is hard to directly verify the sharing surface assumption. Therefore, we use the recent developed panoptic segmentation ground truth to verify if the nearest point is coming from the same object. We believe this evidence can partially support our sharing surface assumption. In Fig. \ref{fig:s_hhh}, all samples are picked from KITTI depth completion validation set. We implement the state-of-the-art panoptic-deeplab as the panoptic segmentation model, and train the model on Cityscape, then fine-tune it on recent released KITTI360. All the black pixels on the depth image are empty pixels that have a nearest point coming from other objects. We can see those pixels are almost all on the edge. This evidence at least shows that most empty pixels share the same object with the nearest point. So we think it partially support our sharing surface assumption.

\subsection{Part 3. More Outlier Mask Visualization}
Here we visualize more samples to show the outlier mask generated by our outlier removal algorithm in Fig .\ref{fig:s_hhhh}. We can see most outlier points are near the boundary between foreground objects and the background.





\bibliographystyle{IEEEtran}
\bibliography{IEEEabrv,IEEEexample}

\end{document}